\documentclass[10pt,conference,a4paper]{IEEEtran}
\IEEEoverridecommandlockouts
\usepackage{cite}
\usepackage{amsmath,amssymb,amsfonts}
\usepackage{algorithmic}
\usepackage{graphicx}
\usepackage{textcomp}
\usepackage{xcolor}
\def\BibTeX{{\rm B\kern-.05em{\sc i\kern-.025em b}\kern-.08em
    T\kern-.1667em\lower.7ex\hbox{E}\kern-.125emX}}

\usepackage{booktabs}
\usepackage{colortbl}
\usepackage{comment}
\usepackage{multirow}
\usepackage{subcaption}
\usepackage{tablefootnote}
\usepackage{tabularx}
\usepackage{url}

\begin{document}

\title{Neural Compression and Filtering for Edge-assisted Real-time Object Detection in Challenged Networks
\thanks{Accepted to the 25th International Conference on Pattern Recognition (ICPR 2020, 1st round). This work was partially supported by DARPA under grant HR00111910001, NSF under Grant IIS-1724331 and MLWiNS-2003237, and Google Cloud Platform research credits.}
}

\author{\IEEEauthorblockN{Yoshitomo Matsubara}
\IEEEauthorblockA{
% \textit{Department of Computer Science} \\
\textit{University of California, Irvine}\\
Irvine, CA \\
yoshitom@uci.edu}
\and
\IEEEauthorblockN{Marco Levorato}
\IEEEauthorblockA{
% \textit{Department of Computer Science} \\
\textit{University of California, Irvine}\\
Irvine, CA \\
levorato@uci.edu}
}

\maketitle

\begin{abstract}
The edge computing paradigm places compute-capable devices - edge servers - at the network edge to assist mobile devices in executing data analysis tasks. Intuitively, offloading compute-intense tasks to edge servers can reduce their execution time. However, poor conditions of the wireless channel connecting the mobile devices to the edge servers may degrade the overall capture-to-output delay achieved by edge offloading. 
Herein, we focus on edge computing supporting remote object detection by means of Deep Neural Networks (DNNs), and develop a framework to reduce the amount of data transmitted over the wireless link. The core idea we propose builds on recent approaches splitting DNNs into sections - namely head and tail models - executed by the mobile device and edge server, respectively. The wireless link, then, is used to transport the output of the last layer of the head model to the edge server, instead of the DNN input. Most prior work focuses on classification tasks and leaves the DNN structure unaltered. Herein, our focus is on DNNs for three different object detection tasks, which present a much more convoluted structure, and modify the architecture of the network to: (i) achieve in-network compression by introducing a bottleneck layer in the early layers on the head model, and (ii) prefilter pictures that do not contain objects of interest using a convolutional neural network. Results show that the proposed technique represents an effective intermediate option between local and edge computing in a parameter region where these extreme point solutions fail to provide satisfactory performance. The code and trained models are available at \textbf{\url{https://github.com/yoshitomo-matsubara/hnd-ghnd-object-detectors}}.
\end{abstract}

\section{Introduction}
The real-time execution of modern data analysis algorithms requires powerful computing platforms. For instance, accurate Deep Neural Networks (DNNs) for vision tasks, \emph{e.g.}, image classification and object detection, have an extremely large number of layers and parameters.
Due to weight, cost or size constraints, mobile devices may have limited computing capacity and energy availability, which makes the execution of such algorithms challenging.
The research community is exploring two main approaches to mitigate this problem: simplifying models and computation offloading.
%\yoshiout{On the one hand, simplified versions of DNNs specifically designed to fit within the capabilities of mobile devices are being developed.
%On the other hand, the possibility to offload compute-intense tasks to servers located at the network edge is being increasingly considered to reduce both energy consumption and overall analysis time.}
The former approach may lead to a degraded performance of the resulting models compared to the full sized ones.
% The degradation is more apparent when model reduction techniques are applied extremely complex tasks such as object detection.
The latter approach is often referred to as \emph{mobile edge computing}, which has been proposed primarily as a tool to support time-sensitive mission-critical applications such as autonomous vehicles and augmented reality~\cite{Gruteser19}.

However, poor channel conditions increase the time needed to deliver the input data to the edge server, thus making edge computing schemes not effective.
Compression, by reducing the channel capacity needed to sustain the data transfer, can mitigate this problem and reduce the total delay.
Unfortunately, while lossless compression is either computationally expensive and/or achieves limited compression gain~\cite{Matsubara19}, traditional lossy compression mechanisms, such as JPEG, are mainly designed for human perception.
As a result, aggressive compression often degrades the final accuracy of analysis~\cite{Grulich18,Xie19}. Modern compression strategies, such as autoencoders~\cite{vincent2008extracting}, are computationally expensive, especially when considering information-rich signals such as images and videos.

Recent contributions~\cite{Kang17,Li18} attempt to distribute the execution of DNN models across the mobile device-edge server system to find delay-optimal ``splitting'' points. More specifically, the DNN model is \emph{split} into a \emph{head} and \emph{tail} model, which are executed at the mobile device and edge server, respectively. Instead of the overall model input, then, the mobile device transmits over the channel the output of the head layer (a tensor). Unfortunately, most of the models, and especially those for vision tasks, tend to amplify the input in the first layers, meaning that positioning the splitting point in the first part of the DNN would result in a larger amount of data transmitted over the wireless link compared to ``pure'' offloading~\cite{Matsubara19}.
The use of later, smaller, layers as splitting point would reduce the amount of transferred data, but would also result in a large portion of the overall computing load being allocated to the weaker device in the system -- the mobile device. In fact, results in~\cite{Kang17} show that in most of the analyzed models and channel/computing settings the optimal splitting point is either at the very beginning or end of the model, that is, pure local computing or edge computing have better performance compared to naive splitting.

%%% not sure I want to talk here about compression

In this paper, we explore strategies to modify the structure of Convolutional Neural Network (CNN) models for object detection to maximize the efficiency of model splitting.
The core idea is to use network distillation to introduce a bottleneck layer, that is, a layer with a small number of nodes, in the early stages of the object detection model. The model is then split at the bottleneck, thus achieving in-network compression, as a smaller tensor needs to be transmitted to the edge server.

We show that the distillation technique we propose allows the injection of extremely effective bottlenecks while modifying only the parameters of the head portion of the model, which is trained to mimic the output of the original head. We build on the results presented in~\cite{Matsubara19}, where we applied a similar concept to image classification tasks.
The more convoluted structure of object detectors, where the output of intermediate layers is propagated to a final detection module, introduces further challenges discussed in~\cite{matsubara2020split} that we address by proposing a generalized loss function guiding the distillation process and applying a bottleneck quantization technique.

Additionally, we observe that while in image classification tasks all images are classified, in object detection tasks only a fraction of the images may contain objects of interest.
To take advantage of this feature of the problem, we embed in the head model a small CNN whose binary output indicates whether or not the image contains objects to be analyzed by the detector.
The network acts as a filter, blocking empty images while promptly returning an empty detection output, thus also reducing channel usage and server load. We demonstrate that this classifier can be efficiently attached to the first layers of the head model minimizing additional complexity by reusing a portion of the detector.

Results show that our distillation based technique achieves detection performance, measured in terms of mean Average Precision (mAP), comparable to that of state-of-the-art models.
In the evaluation, we use established datasets associated with complex detection tasks \emph{e.g.,} person keypoints detection.
We demonstrate that we can significantly reduce the total time from image capture to the availability of the detector's output compared with local and edge computing in a region of parameters where these extreme-point options struggle to provide satisfactory performance.
To ensure reproducibility of the experimental results and facilitate research in this important research problem, we publish all the code and model weights.\footnote{\url{https://github.com/yoshitomo-matsubara/hnd-ghnd-object-detectors}}

%%%%%%%%%%%%%%%%%%%%%%%%%%%%%%%%%%%%%%%%%%%%%%%%%%%%%%%%%%%%%%%%%%%%%
%%%%%%%%%%%%%%%%%%%%%%%%%%%%%%%%%%%%%%%%%%%%%%%%%%%%%%%%%%%%%%%%%%%%%
%%%%%%%%%%%%%%%%%%%%%%%%%%%%%%%%%%%%%%%%%%%%%%%%%%%%%%%%%%%%%%%%%%%%%
%%%%%%%%%%%%%%%%%%%%%%%%%%%%%%%%%%%%%%%%%%%%%%%%%%%%%%%%%%%%%%%%%%%%%
\section{Edge-Assisted Object Detection}
\label{sec:system_desc}

%Figure~\ref{fig:system} depicts the system considered in this paper, 
We consider a scenario where a mobile device acquires images to be analyzed in real-time to support delay-sensitive vision-based applications. An edge server interconnected to the mobile device through a capacity-limited wireless channel (\emph{e.g.}, data rate $\leq 10$Mbps) assists the execution of the analysis algorithm.
As stated earlier, in this paper we focus on object detection tasks.
An example of relevant application and framework is presented in~\cite{Gruteser19}, where the authors focus on an edge-assisted implementation of augmented reality.%, where object detection is used in conjunction with object tracking to improve frame rate. The two algorithms operate at different temporal scales, where only a few frames are analyzed using object detection, whose output is used as a reference for object tracking executed on all frames. In the scenario considered in~\cite{Gruteser19}, object tracking is executed at the mobile device, whereas object detection, which has a much higher computational complexity, is executed at the edge server. 

Our main objective is to reduce the time from image capture to the availability of the object detection output (in this case bounding boxes and associated labels) at the mobile device.
We refer to this time as \emph{capture-to-output delay} and denote it with $T$. Intuitively, in settings such as that proposed in~\cite{Gruteser19}, a smaller capture-to-output delay would improve tracking performance.
We define in the following the composition of $T$ in three settings: (\emph{i}) local computing at the mobile device; (\emph{ii}) offloading of the complete model execution to the edge server (referred to as pure offloading); and (\emph{iii}) split computing, where the execution of the model is divided between the mobile device and the edge server.

\vspace{1mm}
\noindent
{\bf (i) Local Computing:} The total time $T$ is the time needed to execute the entire object detection model at the mobile device. Thus, $T{=}T_{\rm L}$, where $T_{\rm L}$ is determined by the model complexity and computing power of the mobile device.

\vspace{1mm}
\noindent
{\bf (ii) Pure Offloading:} The total time $T$ is the sum of two terms: $T_{\rm i}$ is the time needed to transfer the image to the edge server, and $T_{\rm E}$ is the execution time of the entire model at the edge server.

\vspace{1mm}
\noindent
{\bf (iii) Split Computing:} The total time $T$ is the sum of three terms: $T_{\rm H}$ is the time needed to execute the head model at the mobile device, $T_{\rm o}$ is the time needed to transfer the output of the head model to the edge server, and $T_{\rm T}$ is the execution time of the tail model at the edge server.

\vspace{1mm}
%In the evaluation section we will also consider the time needed to execute the modified head plus tail pipeline at an individual device to provide a comparison with the original model in terms of computational complexity. Moreover, we will also provide results obtained when distilling the entire model into a smaller one to assess accuracy degradation when attempting to fit state of the art models within the computation capabilities of mobile devices.

Rather intuitively, the absolute and relative computing capacity of the mobile device and edge server and the channel capacity determine the value of the delay components listed above, and thus, which option is the most advantageous in terms of inference time.
Clearly, a large channel capacity would decrease communication-related delay components, eventually making pure offloading a preferable option, and increasingly reducing the need for compression.
A small gap in terms of computing capacity between the mobile device and the edge server would make offloading and network splitting effective only in scenarios with high channel capacity. 

We emphasize that none of the above options can be declared best for all parameters.
Instead, in different parameter regions different options will lead to the smallest overall capture-to-output delay. The technique proposed here realizes an intermediate option between local and edge computing where a small portion of the computing load is allocated to the local device to reduce the amount of transferred data.
Clearly, this option is functional in the lower-intermediate range of channel capacity, and when the gap between the mobile device and edge server is not making the allocation of even small computing loads to the mobile device undesirable.

We remark that the latter characteristics is compatible with the scenarios created by the recent advances in miniaturization, which are bringing considerable computing power within the reach of mobile devices.
Examples include the NVIDIA Jetson Nano and TX2 embedded computers. However, as illustrated in our results, even relatively low-end full-sized servers can significantly reduce execution time, thus making offloading meaningful. %Finally, mobility, challenging propagation environments, channel crowding, and limited available energy are all factors that can reduce the perceived data rate and make the proposed approach advantageous with respect to the two extreme points in computing load allocation.

%We note that we consider a setting where images emitted from the mobile device are encoded individually. Clearly, temporal correlation across captured images could be used to reduce the amount of transferred data (\emph{e.g.}, using video encoding). Herein, we assume that subsequent images have low coherence and temporal encoding would not provide a substantial advantage. This assumption can correspond to a setting where images are emitted sporadically, or only when a substantial change in the captured range occurred. In the context of the application scenario described in~\cite{Gruteser19}, a substantial change would impair tracking performance, so that a new reference from object detection would need to be obtained.

%%%%%%%%%%%%%%%%%%%%%%%%%%%%%%%%%%%%%%%%%%%%%%%%%%%%%%%%%%%%%%%%%%%%%
%%%%%%%%%%%%%%%%%%%%%%%%%%%%%%%%%%%%%%%%%%%%%%%%%%%%%%%%%%%%%%%%%%%%%
%%%%%%%%%%%%%%%%%%%%%%%%%%%%%%%%%%%%%%%%%%%%%%%%%%%%%%%%%%%%%%%%%%%%%
%%%%%%%%%%%%%%%%%%%%%%%%%%%%%%%%%%%%%%%%%%%%%%%%%%%%%%%%%%%%%%%%%%%%%
\section{CNN-Based Detection Models}
\label{sec:structure}

In order to better illustrate our proposed approach and the associated challenges, we first discuss the structure of state-of-the-art object detectors based on deep neural networks. CNN-based models have become the mainstream option for object detection tasks~\cite{Girshick14,Girshick15,Redmon16}. These complex object detection models are categorized into two main classes: single-stage or two-stage object detectors.
Single-stage models~\cite{Redmon16,Liu16} are designed to directly predict bounding boxes and the corresponding object classes given an image. Conversely, two-stage models~\cite{Ren15,He17} generate region proposals as output of the first stage, which are then classified in the second stage.
In general, single-stage detection models have lower overall complexity, and thus execution time, compared to two-stage models. However, two-stage models outperforms single-stage ones in terms of detection performance.

\subsection{Benchmark Models}
%\yoshiout{Recent object detectors such as Mask R-CNN and SSD~\\cite{He17,Sandler18} adopt state-of-the-art image classification models, such as ResNet~\\cite{He16} and MobileNet v2, as \emph{backbone}. The main role of backbones in detection tasks is feature extraction, and models pretrained on a large image dataset such as ILSVRC dataset~\\cite{Krizhevsky14}, are often used to extract accurate feature representations from images when training detection models.
%As illustrated in Fig.~\ref{fig:distillation}, such features include outputs of multiple intermediate layers such as those in the backbone. All the features are fed to complex modules specifically designed for detection tasks, such as the feature pyramid network~\\cite{Lin17}, to further extract high-level semantic feature maps at different scales. Finally, these features are used for bounding box regression and object class prediction.
%}

%In computation offloading, sensor devices will not hold models, but an edge server does.
%Once the sensor device captures images, it can send them to the edge server so that can run models fast, and save the inference time.
%However, either knowledge distillation or computation offloading would not be the best option in detection tasks in terms of inference time.

In this study, we focus our attention on state-of-the-art two-stage models in resource-constrained edge computing systems.
Specifically, we consider Faster R-CNN, Mask R-CNN, and Keypoint R-CNN (from torchvision) with ResNet-50 FPN~\cite{Ren15,He17,Paszke19} pretrained on the COCO 2017 datasets.
Due to the limited space, more details of the models and tasks are provided in Appendix~\ref{appendix:benchmark_model}.

\subsection{Motivations}
\label{subsec:motivations}

Clearly, mobile devices are often unable to execute these strong, but rather complex and convoluted, detectors. In order to obtain models within the reach of weak computing platforms, the research community developed an approach known as \emph{knowledge distillation}~\cite{Hinton14}. The core idea of knowledge distillation is to build a lower complexity, ``student'' model, trained to mimic the output of a pretrained, higher-complexity ``teacher'' model. The key assumption is that large -- teacher -- models are often overparameterized, and can be reduced without a significant performance loss. Interestingly, it is widely known that student models trained to mimic the behavior of their teacher models significantly outperform those trained on the original training dataset~\cite{Ba14}.

\begin{table}[tb]
\caption{Local computing time [sec] of detection models with different ResNet backbones on NVIDIA Jetson TX2.}
\vspace{-1em}
\begin{center}
\small
    \begin{tabular}{l|rrrr} \toprule
    \multicolumn{1}{c|}{Model \textbackslash Backbone} & \multicolumn{1}{c}{RN-18} & \multicolumn{1}{c}{RN-34} & \multicolumn{1}{c}{RN-50} & \multicolumn{1}{c}{RN-101} \\ \midrule
    Faster R-CNN & 0.617 & 0.743 & 0.958 & 1.26 \\ 
    Mask R-CNN & 0.645 & 0.784 & 0.956 & 1.27 \\ 
    Keypoint R-CNN & 1.93 & 2.09 & 2.25 & 2.59 \\ \bottomrule
	\end{tabular}
\end{center}
\label{table:local_proc}

\vspace{0.5em}
\caption{Pure offloading time [sec] (data rate: 5Mbps) of detection models with different ResNet backbones on a high-end machine with a NVIDIA GeForce RTX 2080 Ti.}
\vspace{-1em}
\begin{center}
\small
    \begin{tabular}{l|rrrr} \toprule
    \multicolumn{1}{c|}{Model \textbackslash Backbone} & \multicolumn{1}{c}{RN-18} & \multicolumn{1}{c}{RN-34} & \multicolumn{1}{c}{RN-50} & \multicolumn{1}{c}{RN-101} \\ \midrule
    Faster R-CNN & 0.456 & 0.462 & 0.472 & 0.489 \\ 
    Mask R-CNN & 0.458 & 0.4832 & 0.4904 & 0.4897 \\
    Keypoint R-CNN & 0.469 & 0.473 & 0.481 & 0.498 \\ \bottomrule
	\end{tabular}
\end{center}
\label{table:pure_offload}
\end{table}

Distillation has been applied to detection models~\cite{Li17,Chen17,Wang19}.
For instance, Chen \emph{et al.}~\cite{Chen19} propose a hierarchical knowledge distillation technique to train a lightweight pedestrian detector, and distill a student ResNet-18 based R-CNN model using a ResNet-50 based R-CNN model as teacher.
However, as shown in Table~\ref{table:local_proc}, the reduction in inference time granted by smaller models is limited when using relatively capable platforms such as the NVIDIA Jetson TX2, which embeds a GPU.
Importantly, we remark that smaller models achieve degraded detection performance compared to bigger models due to the complexity of the detection task.

Table~\ref{table:pure_offload} shows the total time $T$ achieved by pure offloading when using the same models.
In these results, the execution time is computed using a high-end desktop computer with Intel Core i7-6700K CPU (4.00GHz), 32GB RAM, and a NVIDIA GeForce RTX 2080 Ti as edge server, and the channel provides the relatively low, data rate of $5$Mbps to the image stream.
It can be seen how, in this setting, reducing the model size by distilling the whole detector~\cite{Li17,Chen17,Wang19} does not lead to substantial total delay savings, while offloading is generally advantageous compared to local computing.

Splitting the model while leaving its structure unaltered, as proposed by Kang~\emph{et al.}~\cite{Kang17}, is technically challenging and intuitively not advantageous due to the data amplification effect of early layers, and the need to forward the output of intermediate backbone layers (see Fig.~\ref{fig:distillation}) when positioning the splitting point later in the model.

\section{In-Network Neural Compression}
\label{sec:distill}

\subsection{Background}
As discussed earlier, the weak point of pure edge computing is the communication delay: when the capacity of the channel interconnecting the mobile device and edge server is degraded by a poor propagation environment, mobility, interference and/or traffic load, transferring the model input to the edge server may result in a large overall capture-to-output delay $T$. Thus, in challenged channel conditions, making edge computing effective necessitates strategies to reduce the amount of data to be transported over the channel.

The approach we take herein is to modify the structure of the model to obtain in-network compression and improve the efficiency of network splitting. We remind that in network splitting, the output of the last layer of the head model is transferred to the edge, instead of the model input. Compression, then, corresponds to splitting the model at layers with a small number of nodes which generate small outputs to be transmitted over the channel. In our approach, the splitting point coincides with the layer where compression is achieved.

Unfortunately, layers with a small number of nodes appear only in advanced portions of object detectors, while early layers amplify the input to extract features. However, splitting at late layers would position most of the computational complexity at the weaker mobile device. This issue was recently discussed in~\cite{Matsubara19,matsubara2020split} for image classification and object detection models, reinforcing the results obtained in~\cite{Kang17} on traditional network splitting.

In~\cite{Matsubara19}, we proposed to inject \emph{bottleneck layers}, that is, layers with a small number of nodes, in the early stages of image classification models.
To reduce accuracy loss as well as computational load at the mobile device, the whole head section of the model is reduced using distillation. The resulting small model contains a bottleneck layer followed by a few layers that translate the bottleneck layer's output to the output of the original head model. Note that the layers following the bottleneck layers are then attached to the original tail model and executed at the edge server.

The distillation process attempts to make the output of the new head model as close as possible to the original head. At an intuitive level, when introducing bottleneck layers this approach is roughly equivalent to train a small asymmetric autoencoder-decoder pipeline whose boundary layer produces a compressed version of the input image used by the decoder to reconstruct the output of the head section, rather than the image. Interestingly, it is shown that the distillation approach can achieve high compression rates while preserving classification performance even is complex classification tasks.

This paper builds on this approach~\cite{Matsubara19,matsubara2020split} to obtain in-network compression with further improved detection performance in object detection tasks.
Specifically, we generalize the head network distillation (HND) technique, and apply it to the state of the art detection models described in the previous section (Faster R-CNN, Mask R-CNN, and Keypoint R-CNN).

The key challenge originates from the structural differences between the models for these two classes of vision tasks. As discussed in the previous section, although image classification models are used as backbones of detection models, there is a trend of using outputs of multiple intermediate layers as features fed to modules designed for detection such as the FPN (Feature Pyramid Network)~\cite{Lin17}.
This makes the distillation of head models difficult, as they would need to embed multiple bottleneck layers at the points in the head network whose output is forwarded to the detectors.
Clearly, the amount of data transmitted over the network would be inevitably larger as multiple outputs would need to be transferred. Additionally, we empirically show that injecting smaller bottlenecks resulted in degraded detection performance~\cite{matsubara2020split}.

%Matsubara \emph{et al.} proposed the original head network distillation technique, and introduce small bottleneck to image classification models so that they can split the models at the bottleneck point and minimize the total inference time~\cite{Matsubara19}.

To overcome this issue, we redefine here the head distillation technique to (\emph{i}) introduce the bottleneck at the very early layers of the network, and (\emph{ii}) refine the loss function used to distill the mimicking head model to account for the loss induced on forwarded intermediate layers' outputs. We remark that in network distillation (see Fig.~\ref{fig:distillation}) applied to head-tail split models, the head portion of the model (red) is distilled introducing a bottleneck layer, and the original teacher's architecture and parameters for the tail section (green) are reused without modification. We note that this allows fast training, as only a small portion of the whole model is retrained.

% \vspace{2mm}
% \noindent
% {\bf Bottleneck Positioning and Head Structure - }
\subsection{Bottleneck Positioning and Head Structure}
\begin{figure}[t]
    \centering
    \includegraphics[width=1\linewidth]{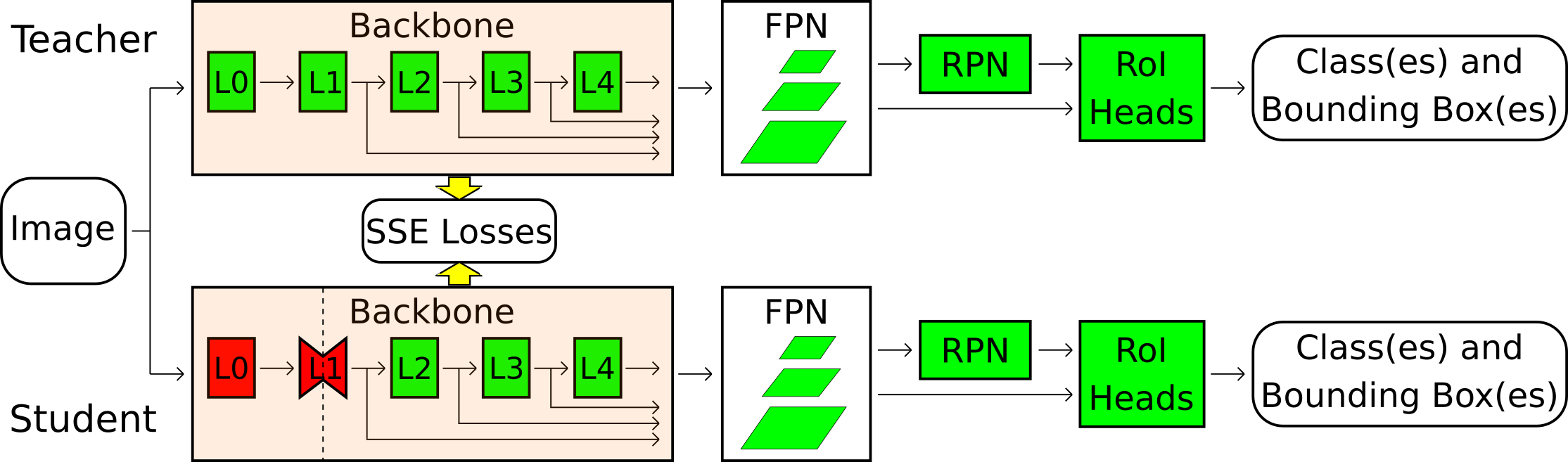}
    \caption{Generalized head network distillation for R-CNN object detectors. Green modules correspond to frozen blocks of individual layers of/from the teacher model, and red modules correspond to blocks we design and train for the student model. L0-4 indicate high-level layers in the backbone. In this work, only backbone modules (orange) are used for training.}
    \label{fig:distillation}
\end{figure}

As shown in Fig.~\ref{fig:distillation}, the output of early blocks of the backbone are forwarded to the detectors. In order to avoid the need to introduce the bottlenecks in multiple sections and transmit their output, we introduce the bottleneck within the L1 module of the model, whose output is the first to be forwarded to FPN. Compared to the framework developed in~\cite{Matsubara19}, this has two main implications. Firstly, the aggregate complexity of the head model is fairly small, and we do not need to significantly reduce its size to minimize computing load at the mobile device. Secondly, in these first layers the extracted features are not yet well defined, and devising an effective structure for the bottleneck is challenging.

Figure~\ref{fig:distillation} summarizes the architecture. The difference between the overall teacher and student models are the high-level layers 0 and 1 (L0 and L1), while the rest of the architecture and their parameters is left unaltered.
The architecture of L0 in the student models is also identical to that in the teacher models, but their parameters are retrained during the distillation process.
The L1 in student models is designed to have the same output shape as the L1 in teacher models, while we introduce a bottleneck layer within the module. 

The architectures of layer 1 in teacher and student models are summarized in Appendix~\ref{appendix:net_arch}.
The architecture will be used in all the considered object detection models: Faster, Mask, and Keypoint R-CNNs.
Our introduced bottleneck point is designed to output a tensor whose size is approximately $6-7$\% of the input one.
Specifically, we introduce the bottleneck point by using an aggressively small number of output channels in the convolution layer and amplifying the output with the following layers.
As we design the student's layers, tuning the number of channels in the convolution layer is a key for our bottleneck injection.

The main reason we consider the number of channels as a key hyperparameter is that different from CNNs for image classification, the input and output tensor shapes of the detection models, including their intermediate layers, are not fixed~\cite{Ren15,He17,Paszke19}.
Thus, it would be difficult to have the output shapes of student model match those of teacher model, that must be met for computing loss values in distillation process described later.
For such models, other hyperparameters such as kernel size $k$, padding $p$, and stride $s$ cannot be changed aggressively while keeping comparable detection performance since they change the output patch size in each channel, and some input elements may be ignored depending on their hyperparameter values.
% \emph{i.e.}, Given different size of input tensors, the layer may return the same size of output tensors.
More details about the background of bottleneck positioning and size are given in~\cite{matsubara2020split}.

% \begin{table*}[t]
% \caption{Performance of pretrained and head-distilled models on COCO 2017 validation datasets* for different tasks.}
% \vspace{-1em}
% \begin{center}
%     \begin{tabular}{l|r|rr|rr} \toprule
%     \multicolumn{1}{c|}{Model with FPN} & \multicolumn{1}{c|}{Faster R-CNN} & \multicolumn{2}{c|}{Mask R-CNN} & \multicolumn{2}{c}{Keypoint R-CNN}\\ \cline{1-1} \cline{2-2} \cline{3-4} \cline{5-6}
%     \multicolumn{1}{c|}{Approach \textbackslash ~Metrics} & \multicolumn{1}{c|}{BBox mAP} & \multicolumn{1}{c}{BBox mAP} & \multicolumn{1}{c|}{Mask mAP} & \multicolumn{1}{c}{BBox mAP} & \multicolumn{1}{c}{Keypoints mAP} \\ \midrule
% 	Pretrained (Teacher) $\dagger$ & 0.370 & 0.379 & 0.346 & 0.546  & 0.650 \\ \midrule
% 	Head network distillation~\cite{Matsubara19,matsubara2020split} & 0.339 & 0.350 & 0.319 & 0.488 & 0.579 \\
% 	\textbf{Ours} & \textbf{0.358} & \textbf{0.370} & \textbf{0.337} & \textbf{0.532} & \textbf{0.634} \\
% 	\textbf{Ours with BQ (16 bits)} & \textbf{0.358} & \textbf{0.370} & \textbf{0.336} & \textbf{0.532} & \textbf{0.634} \\
% 	\textbf{Ours with BQ (8 bits)} & \textbf{0.355} & \textbf{0.369} & \textbf{0.336} & \textbf{0.530} & \textbf{0.628} \\ \bottomrule
% 	\end{tabular}
% \end{center}
% \vspace{-0.5em}
% \small \raggedright *~Test datasets for object and keypoint detection tasks are not publicly available. $\dagger$ \url{https://github.com/pytorch/vision/releases/tag/v0.3.0}
% \label{table:distillation_results}
% \end{table*}

% \vspace{2mm}
% \noindent
% {\bf Loss Function - }
\subsection{Loss Function}

In head network distillation initially applied to image classification~\cite{Matsubara19}, the loss function used to train the student model is defined as
\begin{equation}
    Loss_{org}(X) = \sum_{x \in X} || t(x) - s(x)||^2,
    \label{eq:org_loss}
\end{equation}
where $t(x)$ and $s(x)$ are teacher and student functions of input data $x$ in a batch $X$. The loss function, thus, is simply the sum of squared errors (SSE) between the outputs of last student and teacher layers, and the student model is trained to minimize the loss.
This simple approach produced good results in image classification models.

Due to the convoluted structure of object detection models, the design of the loss function needs to be revisited in order to build effective head models.
As described earlier, the output of multiple intermediate layers in the backbone are used as features to detect objects.
As a consequence, the ``mimicking loss'' at the end of L1 in the student model will be inevitably propagated as tensors are fed forward, and the accumulated loss may degrade the overall detection performance for compressed data size~\cite{matsubara2020split}.

For this reason, we reformulate the loss function as follows:
\begin{equation}
    Loss(X) = \sum_{x \in X} \sum_{j \in J} \lambda_{j} \cdot L_{j}(x, t_{j}, s_{j}),
    \label{eq:gen_loss}
\end{equation}
where ${j}$ is loss index, $\lambda_{j}$ is a scale factor (hyperparameter) associated with loss $L_{j}$, and $t_{j}$ and $s_{j}$ indicate the corresponding subset of teacher and student models (functions of input data $x$) respectively.
The total loss, then, is the sum of $|J|$ weighted losses.
Following Eq.~(\ref{eq:gen_loss}), the previously proposed head network distillation technique~\cite{Matsubara19} can be seen as a special case of our proposed technique.

%Using both the original~\cite{Matsubara19} and our generalized head network distillation techniques, we introduce significantly small bottleneck to detection models so that we can split the model at the bottleneck point, and minimize total inference time, compared to those of the full local processing and full computation offloading.

% \vspace{2mm}
% \noindent
% {\bf Detection Performance Evaluation -}
\subsection{Detection Performance Evaluation}
\begin{figure}[tb]
    \centering
    \includegraphics[width=0.85\linewidth]{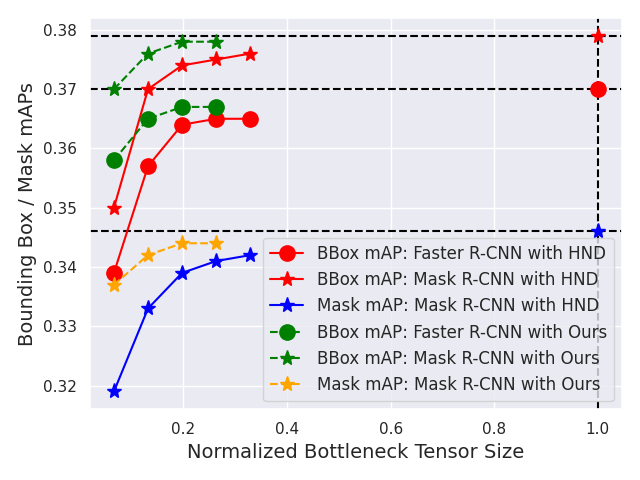}
    \vspace{-1em}
    \caption{Normalized bottleneck tensor size vs. mean average precision of Faster and Mask R-CNNs with FPN.}
    \label{fig:size_vs_map}
\end{figure}

As we modify the structure and parameters of state-of-the-art models to achieve an effective splitting, we need to evaluate the resulting object detection performance. In the following experiments, we use the same distillation configurations for both the original and our generalized head network distillation techniques. Distillations are performed using the COCO 2017 training datasets and the following hyperparameters.
Student models are trained for 20 epochs, and batch size is 4.
The models' parameters are optimized using Adam~\cite{Kingma15} with an initial learning rate of $10^{-3}$, which is decreased by a factor $0.1$ at the 5th and 15th epochs for Faster and Mask R-CNNs.
The number of training samples in the person keypoint dataset is smaller than that in object detection dataset, thus we train Keypoint R-CNN student models for 35 epochs and decrease the learning rate by a factor of 0.1 at the 9th and 27th epochs.

When using the original head network distillation proposed in~\cite{Matsubara19}, the sum of squared error loss is minimized (Eq.~(\ref{eq:org_loss})) using the outputs of the high-level layer 1 (L1) of the teacher and student models.
In the head network distillation for object detection we propose, we minimize the sum of squared error losses in Eq.~(\ref{eq:gen_loss}) using the output of the high-level layers 1--4 (L1--4) with scale factors $\lambda_{*} = 1$. Note that in both the cases, we update only the parameters of the layers 0 and 1, and those of the layers 2, 3 and 4 are fixed.
Quite interestingly, the detection performance degraded when we attempted to update the parameters of layers $1$ to $4$ in our preliminary experiments.
As performance metric, we use mAP (mean average precision) that is averaged over IoU (Intersection-over-Union) thresholds in object detection boxes (BBox), instance segmentation (Mask) and keypoint detection tasks.

\begin{table}[t]
\caption{Performance of pretrained and head-distilled (3ch) models on COCO 2017 validation datasets* for different tasks.}
\vspace{-1em}
\begin{center}
    \bgroup
    \setlength{\tabcolsep}{0.4em}
    \begin{tabular}{l|r|rr|rr} \toprule
    \multicolumn{1}{c|}{R-CNN with FPN} & \multicolumn{1}{c|}{Faster R-CNN} & \multicolumn{2}{c|}{Mask R-CNN} & \multicolumn{2}{c}{Keypoint R-CNN}\\ \cline{1-1} \cline{2-2} \cline{3-4} \cline{5-6}
    \multicolumn{1}{c|}{Approach \textbackslash ~Metrics} & \multicolumn{1}{c|}{BBox} & \multicolumn{1}{c}{BBox} & \multicolumn{1}{c|}{Mask} & \multicolumn{1}{c}{BBox} & \multicolumn{1}{c}{Keypoints} \\ \midrule
	Pretrained (Teacher) $\dagger$ & 0.370 & 0.379 & 0.346 & 0.546  & 0.650 \\ \midrule
	HND~\cite{Matsubara19,matsubara2020split} & 0.339 & 0.350 & 0.319 & 0.488 & 0.579 \\
	\textbf{Ours} & \textbf{0.358} & \textbf{0.370} & \textbf{0.337} & \textbf{0.532} & \textbf{0.634} \\
	\textbf{Ours + BQ (16 bits)} & \textbf{0.358} & \textbf{0.370} & \textbf{0.336} & \textbf{0.532} & \textbf{0.634} \\
	\textbf{Ours + BQ (8 bits)} & \textbf{0.355} & \textbf{0.369} & \textbf{0.336} & \textbf{0.530} & \textbf{0.628} \\ \bottomrule
	\end{tabular}
	\egroup
\end{center}
\vspace{-0.5em}
\footnotesize \raggedright *~Test datasets for these detection tasks are not publicly available.\\
$\dagger$~\url{https://github.com/pytorch/vision/releases/tag/v0.3.0}
\label{table:distillation_results}
\end{table}

Figure~\ref{fig:size_vs_map} reports the detection performance of teacher models, and models with different bottleneck sizes trained by the original and our generalized head network distillation techniques.
For the bottleneck-injected Faster and Mask R-CNNs, the use of our proposed loss function significantly improves mAP compared to models distilled using the original head network distillation (HND)~\cite{matsubara2020split}.
Due to limited space, we show the detection performance of Keypoint R-CNN with an injected bottleneck (3ch) in Table~\ref{table:distillation_results}.
Clearly, the introduction of the bottleneck, and corresponding compression of the output of that section of the network, induces some performance degradation with respect to the original teacher model.

% \vspace{2mm}
% \noindent
% {\bf Bottleneck Quantization (BQ) -}
\subsection{Bottleneck Quantization (BQ)}
Using our generalized head network distillation technique, we introduce small bottlenecks within the student R-CNN models.
Remarkably, the bottlenecks save up to approximately $94$\% of tensor size to be offloaded, compared to input tensor. However, compared to the input JPEG, rather than its tensor representation~\cite{matsubara2020split}, the compression gain is still not satisfactory (see Table~\ref{table:data_size}).
To achieve more aggressive compression gains, we quantize the bottleneck output.
Quantization techniques for deep learning~\cite{Li18,Jacob18} have been recently proposed to \emph{compress} models, reducing the amount of memory used to store them. Here, we instead use quantization to compress the bottleneck output specifically, by representing 32-bit floating-point tensors with 16- or 8-bit.

We can simply cast bottleneck tensors (32-bit by default) to 16-bit, but the data size ratio is still above 1 in Table~\ref{table:data_size}, that means there would be no gain of inference time as it take longer to deliver the data to the edge server compared to pure offloading.
Thus, we apply the quantization technique~\cite{Jacob18} to represent tensors with 8-bit integers and one 32-bit floating-point value. 
Note that quantization is applied after distillation to simplify training.
Inevitably, quantization will result in some information loss, which may affect the detection performance of the distilled models. Quite interestingly, our results indicate that there is no significant detection performance loss for most of the configurations in Table~\ref{table:distillation_results}, while achieving a considerable reduction in terms of data size as shown in Table~\ref{table:data_size}.
In Section~\ref{sec:eval} we report results using 8-bit quantization.

\begin{table}[t]
\caption{Ratios of bottleneck (3ch) data size and tensor shape produced by head portion to input data.}
\vspace{-1em}
\begin{center}
    \begin{tabular}{l|rrrr} \toprule
    \multirow{2}{*}{} & \multicolumn{1}{c}{Input} & \multicolumn{1}{c}{Bottleneck} & \multicolumn{2}{c}{Quantized Bottleneck} \\
    & \multicolumn{1}{c}{(JPEG)} & \multicolumn{1}{c}{32 bits} & \multicolumn{1}{c}{16 bits} & \multicolumn{1}{c}{8 bits} \\ \midrule
    Data size & 1.00 & 2.56 & 1.28 & \textbf{0.643} \\ 
    Tensor shape & 1.00 & \textbf{0.0657} & \textbf{0.0657} & \textbf{0.0657} \\ \bottomrule
	\end{tabular}
\label{table:data_size}
\end{center}
\end{table}

%%%%%%%%%%%%%%%%%%%%%%%%%%%%%%%%%%%%%%%%%%%%%%%%%%%%%%%%%%%%%%%%%%%%%
%%%%%%%%%%%%%%%%%%%%%%%%%%%%%%%%%%%%%%%%%%%%%%%%%%%%%%%%%%%%%%%%%%%%%
%%%%%%%%%%%%%%%%%%%%%%%%%%%%%%%%%%%%%%%%%%%%%%%%%%%%%%%%%%%%%%%%%%%%%
%%%%%%%%%%%%%%%%%%%%%%%%%%%%%%%%%%%%%%%%%%%%%%%%%%%%%%%%%%%%%%%%%%%%%
\section{Neural Image Prefiltering}
\label{sec:neuralfilter}

\begin{figure}[tb]
    \centering
    \begin{subfigure}[b]{0.225\linewidth}
        \centering
        \includegraphics[width=\linewidth]{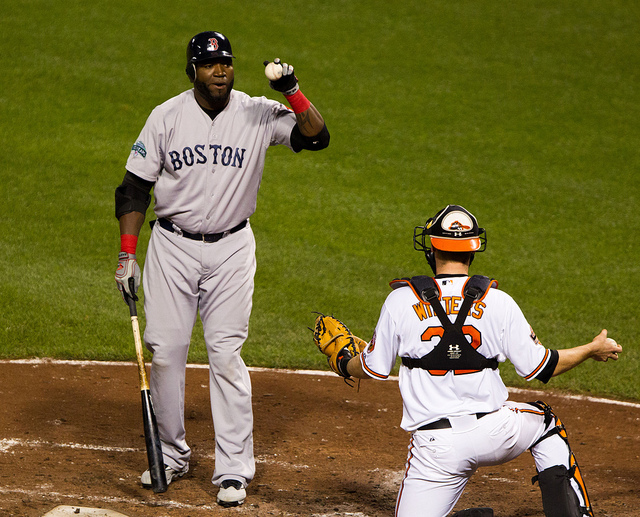}
        \caption{2 persons}
        \label{fig:sample1}
    \end{subfigure}
    \begin{subfigure}[b]{0.24\linewidth}
        \centering 
        \includegraphics[width=\linewidth]{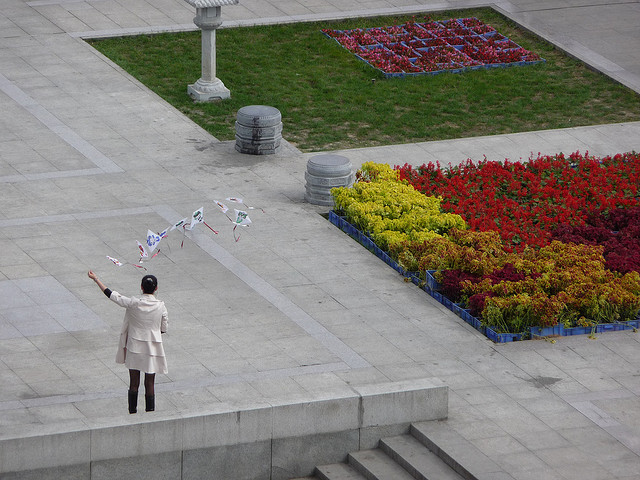}
        \caption{1 person}
        \label{fig:sample2}
    \end{subfigure}
    \begin{subfigure}[b]{0.24\linewidth}
        \centering 
        \includegraphics[width=\linewidth]{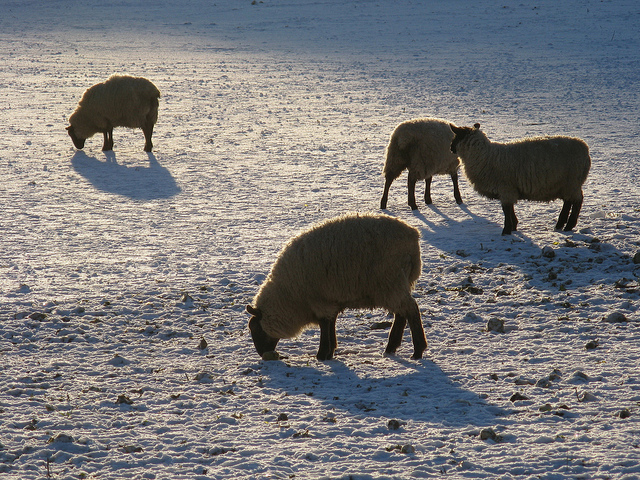}
        \caption{No person}
        \label{fig:sample3}
    \end{subfigure}
    \begin{subfigure}[b]{0.24\linewidth}
        \centering 
        \includegraphics[width=\linewidth]{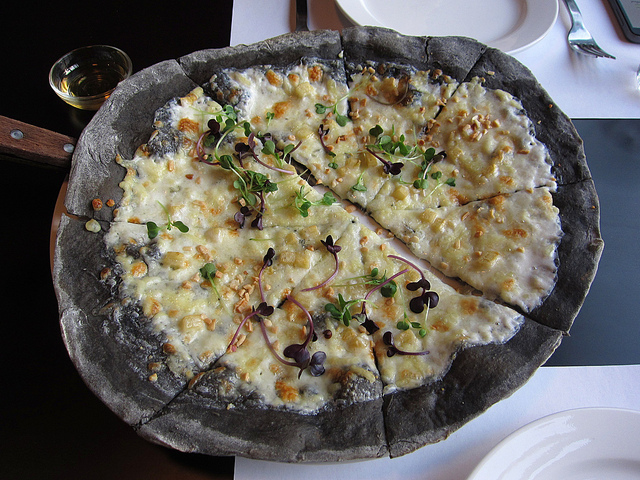}
        \caption{No person}
        \label{fig:sample4}
    \end{subfigure}
    \vspace{-0.5em}
    \caption{Sample images in COCO 2017 training dataset.} 
    \label{fig:sample_images}
\end{figure}

In this section, we exploit a semantic difference between image classification and object detection tasks to reduce resource usage. While every image is used for inference, only a subset of images produced by the mobile device contain objects within the overall set of detected classes. Intuitively, the execution of the object detection module is useful only if at least one object of interest appear in the vision range.
Figures~\ref{fig:sample3} and \ref{fig:sample4} are examples of pictures without objects of interest for Keypoint R-CNN, as this model is trained to detect people and simultaneously locate their keypoints.
We attempt then, to \emph{filter out} the \emph{empty images} before they are transmitted over the channel and processed by the edge server. 

To this aim, we embed in the early layers of the overall object detection model a classifier whose output indicates whether or not the picture is empty. We refer to this classifier as \emph{neural filter}.
Importantly, this additional capability impacts several metrics: (\emph{i}) reduced total inference time, as the early decision as an empty image is equivalent to the detector's output; (\emph{ii}) reduced channel usage, as empty images are eliminated in the head model; (\emph{iii}) reduced server load, as the tail model is not executed when the image is filtered out.

Clearly, the challenge is developing a low-complexity, but accurate classifier. In fact, a complex classifier would increase the execution time of the head portion at the mobile device, possibly offsetting the benefit of producing early \emph{empty} detection. On the other hand, an inaccurate classifier would either decrease the overall detection performance filtering out non-empty pictures, or failing to provide its full potential benefit by propagating empty pictures.
% Importantly powerful object detection models are much more complex compared to models for image classification due to inherent difficulty of the former task: not only object label(s), but also coordinate(s).
% (such as bounding box regression).
% Therefore, it is possible to develop accurate compact classification models with a limited impact on the overall complexity of a full-sized object detector.

\begin{figure}[tb]
    \centering
    \includegraphics[width=0.75\linewidth]{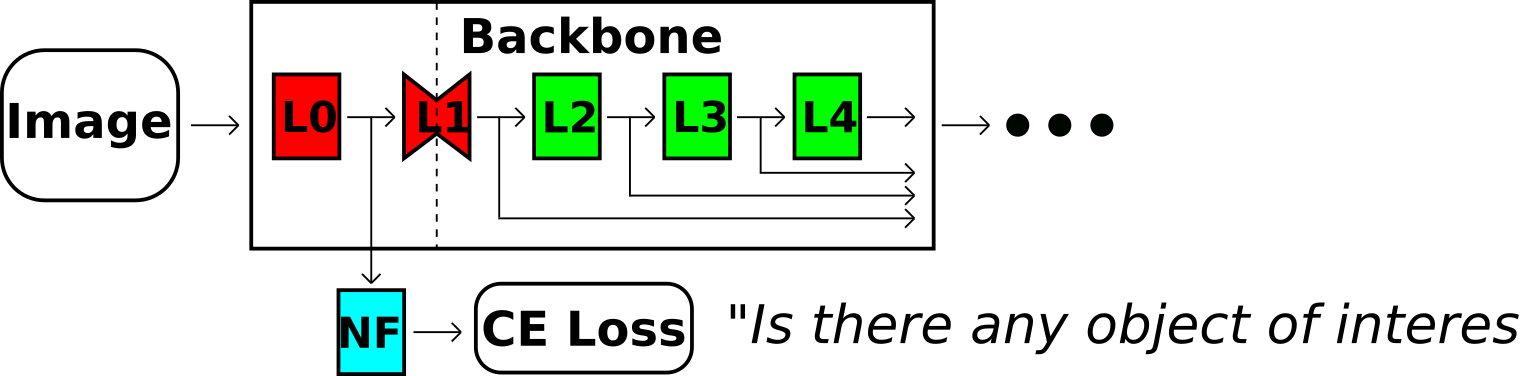}
    \caption{Neural filter (blue) to filter images with no object of interest. Only neural filter's parameters can be updated.}
    \label{fig:neural_filter}
\end{figure}

In the structure we developed in the previous section, we have the additional challenge that the neural filter will need to be attached to the head model, which only contains early layers of the overall detection model (see Fig.~\ref{fig:neural_filter}). Note that parameters of the distilled model are fixed, and only the neural filter (blue module in Fig.~\ref{fig:neural_filter}) is trained. Specifically, as input to the neural filter we use the output of layer L0, the first section of the backbone network. This allows us to reuse layers that are executed in case the picture contains objects to support detection. 
Importantly, the L0 layer performs an amplification of the input data~\cite{He16}. Therefore, using L0 for both of the head model and the neural filter is efficient and effective.

In this study, we introduce a neural filter to a distilled Keypoint R-CNN model as illustrated in Fig.~\ref{fig:neural_filter}.
Approximately 46\% of images in the COCO 2017 person keypoint dataset have no object of interest, and Figures~\ref{fig:sample3} and \ref{fig:sample4} are sample images we would like to filter out.
The design of the neural filter is reported in Appendix~\ref{appendix:net_arch}, and we train the model for $30$ epochs.
Each image is labeled as ``positive'' if it contains at least one valid object, and as ``negative`` otherwise.
We use cross entropy loss to optimize model's parameters by SGD with an initial learning rate $10^{-3}$, momentum $0.9$, weight decay $10^{-4}$, and batch size of $2$. The learning rate is decreased by a factor of 0.1 at the 15th and 25th epochs.
Our neural filter achieved $0.919$ ROC-AUC on the validation dataset.

The output values of the neural filter are softmaxed \emph{i.e.}, $[0,1]$. In order to preserve the performance of the distilled Keypoint R-CNN model when using the neural filter, we set a small threshold for prefiltering to obtain a high recall, while images without objects of interest are prefiltered only when the neural filter is negatively confident.
Specifically, we filter out images with prediction score smaller than $0.1$.
The BBox and Keypoint mAPs of distilled Keypoint R-CNN with BQ (8-bit) and neural filter are $0.513$ and $0.619$ respectively.
As shown in the next section, the detection performance slightly degraded by the neural filter results in a perceivable reduction of the total inference time in the considered datasets.
% Thus, the neural filter only slightly degrades detection performance. However, as shown in the next section, suffering such small degradation results in a perceivable reduction of the total inference time in the considered datasets.

%%%%%%%%%%%%%%%%%%%%%%%%%%%%%%%%%%%%%%%%%%%%%%%%%%%%%%%%%%%%%%%%%%%%%
%%%%%%%%%%%%%%%%%%%%%%%%%%%%%%%%%%%%%%%%%%%%%%%%%%%%%%%%%%%%%%%%%%%%%
%%%%%%%%%%%%%%%%%%%%%%%%%%%%%%%%%%%%%%%%%%%%%%%%%%%%%%%%%%%%%%%%%%%%%
%%%%%%%%%%%%%%%%%%%%%%%%%%%%%%%%%%%%%%%%%%%%%%%%%%%%%%%%%%%%%%%%%%%%%
\section{Latency Evaluation}
\label{sec:eval}

% \begin{table}[tb]
% \caption{Hardware specifications}
% \vspace{-0.5em}
% \begin{center}
% \bgroup
% \setlength{\tabcolsep}{0.3em}
% \def\arraystretch{1.1}
% \footnotesize
% 	\begin{tabular}{llr} \toprule
% 	\multicolumn{1}{c}{Computer} & \multicolumn{1}{c}{Processor} & \multicolumn{1}{c}{RAM} \\ \midrule
% 	\shortstack[l]{Jetson TX2\vspace{0.5em}} & \shortstack[l]{ARM Cortex-A57 (4 cores)\\\hspace{0.1em} + NVIDIA Denver2 (2 cores)} & \shortstack[r]{8GB\vspace{0.5em}} \\
% 	\shortstack[l]{Desktop\vspace{0.5em}} & \shortstack[l]{Intel Core i7-6700K (4 cores)\\\hspace{0.1em} + NVIDIA GeForce RTX 2080 Ti} & \shortstack[r]{32GB\vspace{0.5em}} \\
% 	\bottomrule
% 	\end{tabular}
% \egroup
% \end{center}
% \label{table:hardware}
% \end{table}

In this section, we evaluate the total time $T$ of capture-to-output pipelines.
Following Section~\ref{subsec:motivations}, we use the NVIDIA Jetson TX2 as mobile device, and the high-end desktop computer with a NVIDIA GeForce RTX 2080 Ti as edge server.
% This configuration corresponds to a scenario with relatively capable devices.
Clearly, scenarios with weaker mobile devices and edge servers will see a reduced relative weight of the communication component of the total delay, thus possibly advantaging our technique compared to pure offloading.
On the other hand, a strongly asymmetric system, where the mobile device has a considerably smaller computing capacity compared to the edge server will penalize the execution of even small computing tasks at the mobile device, as prescribed by our approach.

We compare three different configurations: local computing, pure offloading, and split computing using network distillation.
Here, we do not consider naive splitting approaches such as Neurosurgeon~\cite{Kang17} as the original R-CNN models used in this study do not have any small bottleneck point~\cite{matsubara2020split}, and the best splitting point would result in either input or output layers \emph{i.e.}, pure offloading or local computing.
However, we consider the same data rates for vision task as in \cite{Kang17}, and focus thus on rates below $10$Mbps.
Note that all the R-CNN models are designed to have an input image whose shorter side has $800$ pixels.
In pure offloading, we compute the file size in bytes of the resized JPEG images to be transmitted to the edge server. The average size of resized images in COCO 2017 validation dataset is $874 \times 1044$.
In the split configuration, we compute data size of quantized output of the bottleneck layer, and the communication delay is computed dividing the data size by the available data rate.

\begin{figure}[tb]
        \centering
        \begin{subfigure}[b]{0.49\linewidth}
            \centering
            \includegraphics[width=\linewidth]{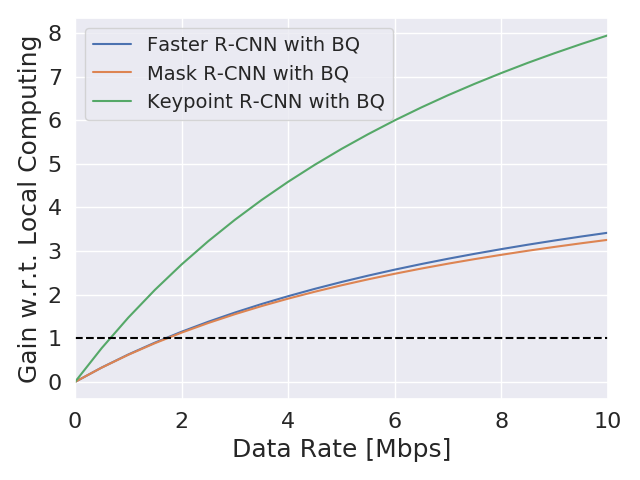}
            \caption{Gain w.r.t. local computing.}  
            \label{fig:gain_wrt_local_comp}
        \end{subfigure}
        \begin{subfigure}[b]{0.49\linewidth}
            \centering 
            \includegraphics[width=\linewidth]{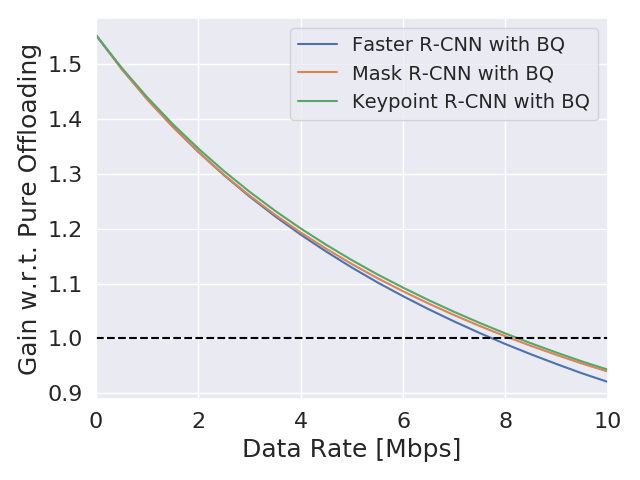}
            \caption{Gain w.r.t. pure offloading.}  
            \label{fig:gain_wrt_pure_offload}
        \end{subfigure}
        \begin{subfigure}[b]{0.49\linewidth}
            \centering 
            \includegraphics[width=\linewidth]{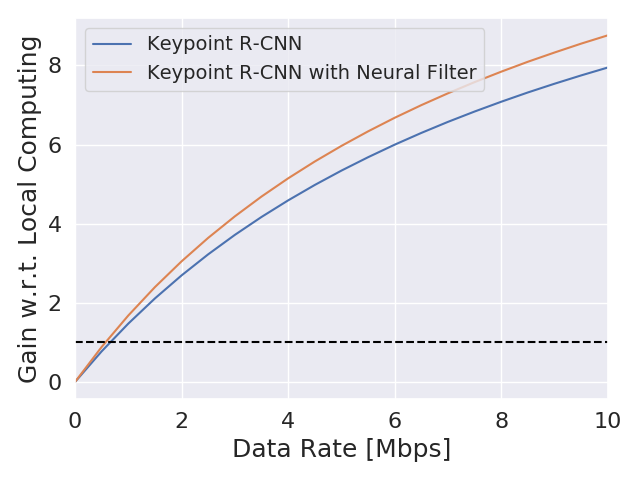}
            \caption{Gain with a neural filter w.r.t. local computing.}    
            \label{fig:ext_gain_wrt_local_comp}
        \end{subfigure}
        \begin{subfigure}[b]{0.49\linewidth}
            \centering 
            \includegraphics[width=\linewidth]{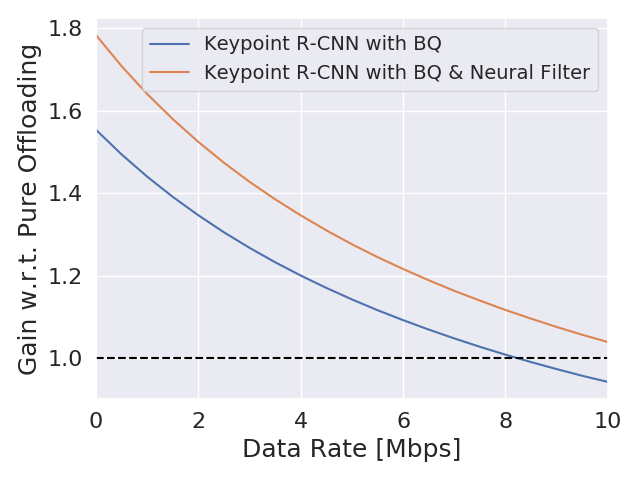}
            \caption{Gain with a neural filter w.r.t. pure offloading.}     
            \label{fig:ext_gain_wrt_pure_offload}
        \end{subfigure}
        \vspace{-1.5em}
        \caption{Ratio of the total capture-to-output time $T$ of local computing and pure offloading to that of the proposed technique without (top)/with (bottom) a neural filter.} 
        \label{fig:gain}
\end{figure}

Figures~\ref{fig:gain_wrt_local_comp} and \ref{fig:gain_wrt_pure_offload} show the gain of the proposed technique with respect to local computing and pure offloading respectively as a function of the available data rate.
The gain is defined as the total delay of local computing/pure offloading divided by that of the split computing configuration.
As expected, local computing is the best option (gain smaller than one) when the available data rate is small. Depending on the specific model, the threshold is hit in the range $0.5-2$Mbps. The gain then grows up to $3$ (Faster and Mask R-CNNs) and $8$ (Keypoint R-CNN) when the data rate is equal to $10$Mbps.

The gain with respect to pure offloading has the opposite trend.
For extremely poor channels, the gain reaches $1.5$, and decreases as the data rate increases until the threshold $1$ is hit at about $8$Mbps.
As we stated in Section~\ref{sec:system_desc}, the technique we developed provides an effective intermediate option between local computing and pure offloading, where our objective is to make the tradeoff between computation load at the mobile device and transmitted data as efficient as possible. 
Intuitively, in this context, naive splitting is suboptimal in any parameter configuration, as the original R-CNN models have no effective bottlenecks for reducing the capture-to-output delay~\cite{matsubara2020split}.
% Intuitively, in this context, naive splitting is suboptimal in any parameter configuration, as the larger time required to execute the (unaltered) head section at the mobile device compared to its execution at the server will offsets the moderate compression rate obtained only at the last backbone layers.
Our technique is a useful tool in challenged networks where many devices contend for the channel resource, or the characteristics of the environment reduce the overall capacity, \emph{e.g.}, non-line of sight propagation, extreme mobility, long-range links, and low-power/low complexity radio transceivers.

%\yoshiout{As a side observation, we note that Keypoint R-CNN emphasizes the gain with respect to local computing. This is due to the larger number of non-Cuda operations in the final prediction part of the model. The smaller RAM of the NVIDIA Jetson TX2 compared to that of the server GPU makes the execution of these operations less efficient and introduces a further source of delay.}

Figures~\ref{fig:ext_gain_wrt_local_comp} and~\ref{fig:ext_gain_wrt_pure_offload} show the same metric when the neural filter is introduced.
In this case, when the neural filter predicts that the input pictures do not contain any object of interest, the tail model on an edge server are not executed. \emph{i.e.}, the system does not offload the rest of computing for such inputs, thus experiencing a lower delay.
The effect is an extension of the data rate ranges in which the proposed technique is the best option, as well as a larger gain for some models. We remark that the results are computed using a specific dataset. Clearly, in this case the inference time is influenced by the ratio of empty pictures. The extreme point where all pictures contain objects collapses the gain to a slightly degraded - due to the larger computing load at the mobile device - version of the configuration without classifier. As the ratio of empty pictures increases, the classifier will provide increasingly larger gains.

% \begin{figure}[tb]
%     \centering
%     \includegraphics[width=1.0\linewidth]{fig/result/inference_time/faster_rcnn_inference_time.png}
%     \caption{Component-wise processing time of Faster R-CNN for different data rates.}
%     \label{fig:faster_rcnn_inference_time}
% \end{figure}

% \begin{figure}[tb]
%     \centering
%     \includegraphics[width=0.9\linewidth]{fig/result/inference_time/mask_rcnn_inference_time.png}
%     \caption{Component-wise processing time of Mask R-CNN for different data rates.}
%     \label{fig:mask_rcnn_inference_time}
% \end{figure}

\begin{figure}[tb]
    \centering
    \includegraphics[width=0.8\linewidth]{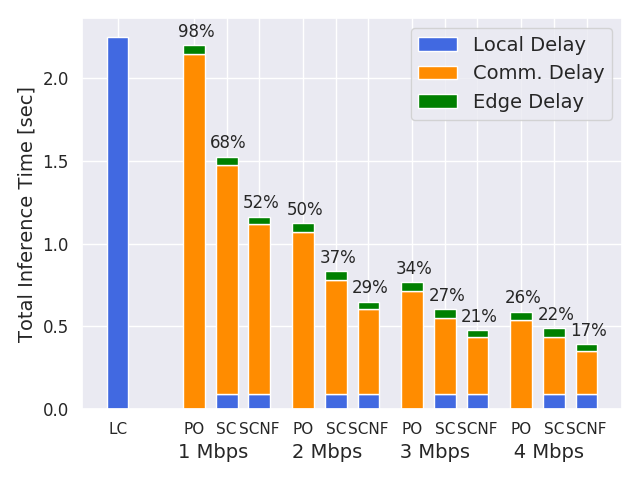}
    \vspace{-1em}
    \caption{Component-wise delays of original and our Keypoint R-CNNs in different data rates. LC: Local Comp., PO: Pure Offload., SC: Split Comp., SCNF: Split Comp. w/ Neural Filter}
    \label{fig:keypoint_rcnn_inference_time}
\end{figure}

We report and analyze the absolute value of the capture-to-output delay for different configurations. Figure~\ref{fig:keypoint_rcnn_inference_time} shows the components of the delay $T$ as a function of the data rate when Keypoint R-CNN is the underlying detector. It can be seen how the communication delays $T_{\rm i}$ (JPEG image) and $T_{\rm o}$ tend to dominate with respect to computing components in the range where our technique is advantageous.
The split approach introduces the local computing component $T_{\rm H}$ associated with the execution of the head portion.
Note that the difference between the execution of the tail portion ($T_{\rm T}$) and the execution of the full model at the edge server ($T_{\rm E}$) is negligible, due to the small size of the head model and the large computing capacity of the edge server.
% The execution time of the head model and the head model plus neural filter are almost equal.
The figure also shows, while the reduction in the total capture-to-output delay is perceivable, the extra classifier imposes a small additional computing load compared to the head model with bottleneck.

%\yoshiout{Figure~\ref{fig:keypoint_rcnn_inference_time} shows the same metrics in the Keypoint R-CNN case, for which we developed the neural classifier.}

%%%%%%%%%%%%%%%%%%%%%%%%%%%%%%%%%%%%%%%%%%%%%%%%%%%%%%%%%%%%%%%%%%%%%
%%%%%%%%%%%%%%%%%%%%%%%%%%%%%%%%%%%%%%%%%%%%%%%%%%%%%%%%%%%%%%%%%%%%%
%%%%%%%%%%%%%%%%%%%%%%%%%%%%%%%%%%%%%%%%%%%%%%%%%%%%%%%%%%%%%%%%%%%%%
%%%%%%%%%%%%%%%%%%%%%%%%%%%%%%%%%%%%%%%%%%%%%%%%%%%%%%%%%%%%%%%%%%%%%
\section{Conclusions}
\label{sec:concl}
Building on recent contributions proposing to split DNNs in head and tail sections~\cite{Matsubara19} executed at the mobile device and edge server respectively, this paper presents a technique to efficiently split deep neural networks for object detection.
The core idea is to achieve in-network compression introducing a bottleneck layer in the early stages of the backbone network.
The output of the bottleneck is quantized and then sent to the edge server, which executes some layers to reconstruct the original output of head model and the tail portion.
Additionally, we embed in the head model a low-complexity classifier which acts as a filter to eliminate pictures that do not contain objects of interest, further improving efficiency.
We demonstrate that our generalized head network distillation can lead to models achieving state-of-the-art performance while reducing total inference time in parameter regions where local and edge computing provide unsatisfactory performance.

\bibliographystyle{IEEEtran}
\bibliography{references}

% \clearpage

\appendices
\renewcommand{\thetable}{A-\Roman{table}}
\renewcommand{\thefigure}{A-\arabic{figure}}

\section{Benchmark R-CNNs}
\label{appendix:benchmark_model}
Faster R-CNN is the strong basis of several 1st-place entries~\cite{He16} in ILSVRC and COCO 2015 competitions.
The model is extended to Mask R-CNN by adding a branch for predicting an object mask in parallel with the existing branch for bounding box recognition~\cite{He17}.
Mask R-CNN is not only a strong benchmark, but also a well-designed framework, as it easily generalizes to other tasks such as instance segmentation and person keypoint detection.

In fact, Mask R-CNN was extended and trained for person keypoint detection (Keypoint R-CNN) in torchvision~\cite{Paszke19}.
In instance segmentation task, models need to precisely predict which pixels belong to which instance while detecting objects in an image.
The keypoint detection task requires the estimation of human poses by simultaneously detecting persons and their keypoints (\emph{e.g.}, joints and eyes)~\cite{Bourdev09} in the image.

\section{Network Architectures}
\label{appendix:net_arch}
Table~\ref{table:teacher_student_arch} reports the network architectures of Layer 1 (L1) in the teacher and student models.
Recall that all the teacher and student models in this study use the architectures of L1 shown in the table. The rest of the models' architecture is not described as, other than L0 and L1, student models have exactly the same architectures as their teacher models.
In the inference time evaluation, we split the student model at the bottleneck layer, (\textbf{bold layer}), to obtain head and tail models, that are executed on the mobile device and edge server, respectively.
The head model consists of all the layers before and including the bottleneck layer, and the remaining layers are used as the corresponding tail model.
We note that in addition to models with the introduced bottleneck used in this study that has 3 output channels (see Table~\ref{table:distillation_results}), 6, 9 and 12 output channels are used (see Fig.~\ref{fig:size_vs_map}) for the bottleneck introduced to exactly the same architectures in \cite{matsubara2020split}.
Such configurations, however, are not considered in this study as the ratios of the corresponding data sizes would be above 1 even with bottleneck quantization, that would result in further delayed inference, compared to pure offloading.

Similarly, Table \ref{table:nf_arch} summarizes the network architecture of the neural filter, where the output of L0 in the ``frozen'' student model is fed to the neural filter.

\begin{table}[h]
\caption{Architectures of Layer 1 (L1) in teacher and student models}
\begin{center}
\bgroup
\setlength{\tabcolsep}{0.3em}
\def\arraystretch{1.1}
% \small
	\begin{tabular}{l|l} \toprule
	\multicolumn{1}{c}{Teacher's L1} & \multicolumn{1}{c}{Student's L1}\\ \midrule
Conv2d(oc=64, k=1, s=1) & Conv2d(oc=64, k=2, p=1) \\
BatchNorm2d & BatchNorm2d \\
Conv2d(oc=64, k=3, s=1, p=1) & Conv2d(oc=256, k=2, p=1) \\
BatchNorm2d & BatchNorm2d \\
Conv2d(oc=256, k=1, s=1) & ReLU \\
BatchNorm2d & Conv2d(oc=64, k=2, p=1) \\
Conv2d(oc=256, k=1, s=1) & BatchNorm2d \\
BatchNorm2d & \textbf{Conv2d(oc=3, k=2, p=1)} \\
ReLU & BatchNorm2d \\
Conv2d(oc=64, k=1, s=1) & ReLU \\
BatchNorm2d & Conv2d(64, k=2) \\
Conv2d(oc=64, k=3, s=1, p=1) & BatchNorm2d \\
BatchNorm2d & Conv2d(oc=128, k=2) \\
Conv2d(oc=256, k=1, s=1) & BatchNorm2d(f=128) \\
BatchNorm2d & ReLU \\
ReLU & Conv2d(oc=256, k=2) \\
Conv2d(oc=64, k=1, s=1) & BatchNorm2d \\
BatchNorm2d & Conv2d(oc=256, k=2) \\
Conv2d(oc=64, k=3, s=1, p=1) & BatchNorm2d \\
BatchNorm2d & ReLU \\
Conv2d(oc=256, k=1, s=1) & \\
BatchNorm2d & \\
ReLU & \\

\bottomrule
	\end{tabular}
\egroup
\end{center}
\small
\raggedright oc: output channel, k: kernel size, s: stride, p: padding. \textbf{A bold layer} is our introduced bottleneck.
\label{table:teacher_student_arch}
\vspace{0.5em}
\caption{Architecture of neural filter}
\vspace{-1em}
\begin{center}
\bgroup
\setlength{\tabcolsep}{0.3em}
\def\arraystretch{1.1}
	\begin{tabular}{l} \toprule
	\multicolumn{1}{c}{Neural filter} \\ \midrule
    AdaptiveAvgPool2d(oh=64, ow=64), \\
    Conv2d(oc=64, k=4, s=2), BatchNorm2d, ReLU, \\
    Conv2d(oc=32, k=3, s=2), BatchNorm2d, ReLU, \\
    Conv2d(oc=16, k=2, s=1), BatchNorm2d, ReLU, \\
    AdaptiveAvgPool2d(oh=8, ow=8), \\
    Linear(in=1024, on=2), Softmax \\
    % AdaptiveAvgPool2d(oh=64, ow=64) \\
    % Conv2d(oc=64, k=4, s=2) \\
    % BatchNorm2d \\
    % ReLU \\
    % Conv2d(oc=32, k=3, s=2) \\
    % BatchNorm2d \\
    % ReLU \\
    % Conv2d(oc=16, k=2, s=1) \\
    % BatchNorm2d \\
    % ReLU \\
    % AdaptiveAvgPool2d(oh=8, ow=8) \\
    % Linear(in=1024, on=2) \\
    % Softmax \\
    \bottomrule
	\end{tabular}
\egroup
\end{center}
\footnotesize
\raggedright oh: output height, ow: output width, in: input feature, on: output feature
\label{table:nf_arch}
\end{table}

% \begin{figure}[tb]
%         \centering
%         \begin{subfigure}[b]{0.49\linewidth}
%             \centering
%             \includegraphics[width=\linewidth]{fig/pic/000000039777.jpg}
%             \caption{Input image\\~}
%             \label{fig:input_image}
%         \end{subfigure}
%         \begin{subfigure}[b]{0.49\linewidth}
%             \centering 
%             \includegraphics[width=\linewidth]{fig/pic/output-faster_rcnn.png}
%             \caption{Output of Faster R-CNN \\(object detection)}
%             \label{fig:faster_rcnn_output}
%         \end{subfigure}
%         \begin{subfigure}[b]{0.49\linewidth}
%             \centering 
%             \includegraphics[width=\linewidth]{fig/pic/output-mask_rcnn.png}
%             \caption{Output of Mask R-CNN \\(object detection and instance segmentation)}
%             \label{fig:mask_rcnn_output}
%         \end{subfigure}
%         \begin{subfigure}[b]{0.49\linewidth}
%             \centering 
%             \includegraphics[width=\linewidth]{fig/pic/output-keypoint_rcnn.png}
%             \caption{Output of Keypoint R-CNN \\(object and person keypoint detections)}
%             \label{fig:keypoint_rcnn_output}
%         \end{subfigure}
%         \vspace{-0.5em}
%         \caption{Input and output generated by different R-CNN models.} 
%         \label{fig:sample_input_outputs}
% \end{figure}

\section{Qualitative Analysis}
\label{appendix:qual_analysis}
\begin{figure*}[h]
    \begin{center}
        \begin{subfigure}[c]{0.24\linewidth}
            \centering
            \includegraphics[width=1.0\linewidth]{fig/pic/input/000000463690.jpg}
            \vspace{-1.5em}
            \caption{Sample input 1 \label{fig:input1}}
        \end{subfigure}
        \begin{subfigure}[c]{0.24\linewidth}
            \centering
            \includegraphics[width=1.0\linewidth]{fig/pic/input/000000493442.jpg}
            \vspace{-1.5em}
            \caption{Sample input 2 \label{fig:input2}}
        \end{subfigure}
        \begin{subfigure}[c]{0.23\linewidth}
            \centering
            \includegraphics[width=1.0\linewidth]{fig/pic/input/000000079144.jpg}
            \vspace{-1.5em}
            \caption{Sample input 3 \label{fig:input3}}
        \end{subfigure}
        \begin{subfigure}[c]{0.24\linewidth}
            \centering
            \includegraphics[width=1.0\linewidth]{fig/pic/input/000000568981.jpg}
            \vspace{-1.5em}
            \caption{Sample input 4 \label{fig:input4}}
        \end{subfigure}
        
        \begin{subfigure}[c]{0.24\linewidth}
            \centering
            \includegraphics[width=1.0\linewidth]{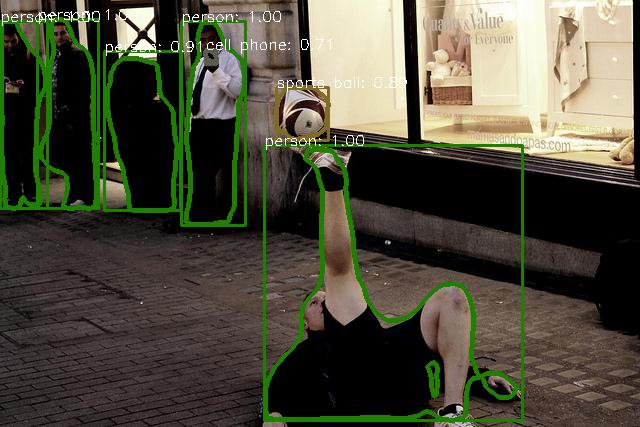}
            \vspace{-1.5em}
            \caption{Mask R-CNN: 5 persons, 1 sports ball, and 1 cell phone\label{fig:mask_org1}}
        \end{subfigure}
        \begin{subfigure}[c]{0.24\linewidth}
            \centering
            \includegraphics[width=1.0\linewidth]{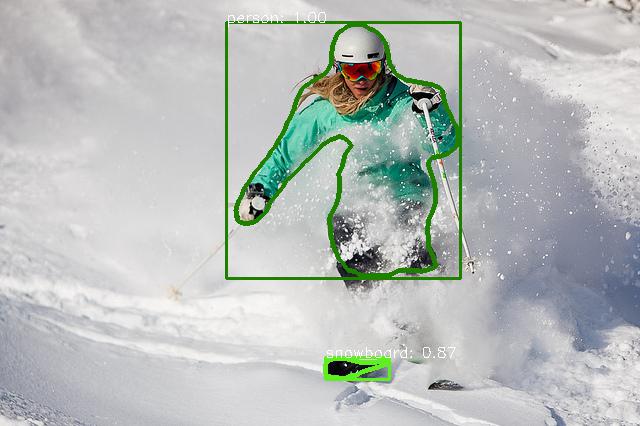}
            \vspace{-1.5em}
            \caption{Mask R-CNN: 1 person and 1 snowboard \label{fig:mask_org2}}
        \end{subfigure}
        \begin{subfigure}[c]{0.23\linewidth}
            \centering
            \includegraphics[width=1.0\linewidth]{fig/pic/keypoint_rcnn/org_000000079144.jpg}
            \vspace{-1.5em}
            \caption{Keypoint R-CNN: No object of interests\label{fig:kp_org1}}
        \end{subfigure}
        \begin{subfigure}[c]{0.24\linewidth}
            \centering
            \includegraphics[width=1.0\linewidth]{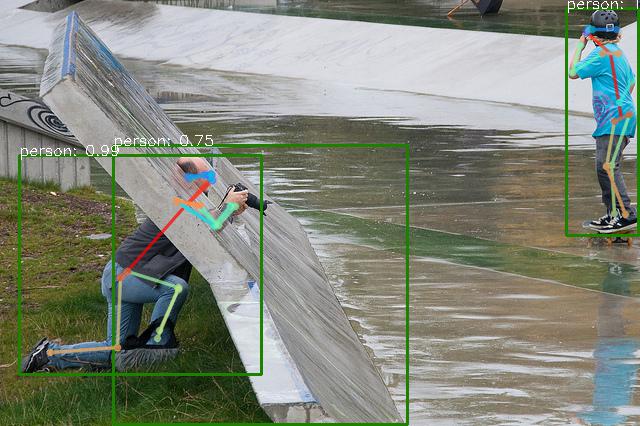}
            \vspace{-1.5em}
            \caption{Keypoint R-CNN: 3 persons\\~\label{fig:kp_org2}}
        \end{subfigure}
        
        \begin{subfigure}[c]{0.24\linewidth}
            \centering
            \includegraphics[width=1.0\linewidth]{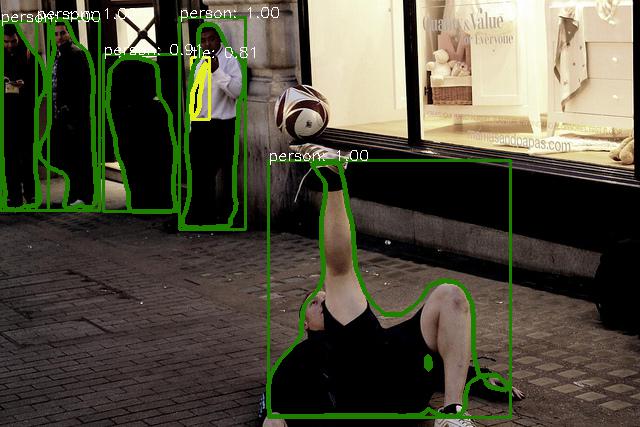}
            \vspace{-1.5em}
            \caption{Our Mask R-CNN distilled by HND~\cite{Matsubara19}: 5 persons and 1 backpack \label{fig:mask_hnd1}}
        \end{subfigure}
        \begin{subfigure}[c]{0.24\linewidth}
            \centering
            \includegraphics[width=1.0\linewidth]{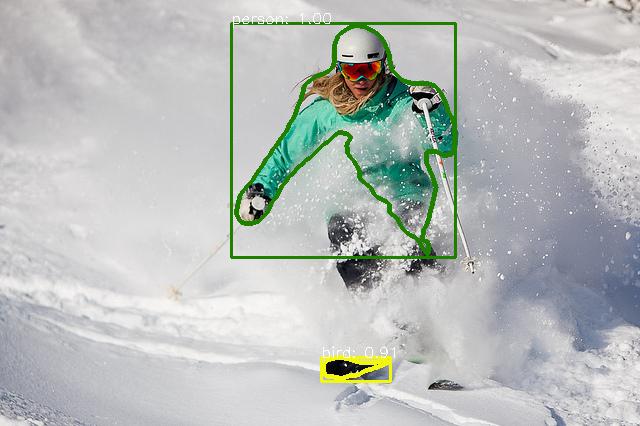}
            \vspace{-1.5em}
            \caption{Our Mask R-CNN distilled by HND~\cite{Matsubara19}: 1 person and 1 bird \\~ \label{fig:mask_hnd2}}
        \end{subfigure}
        \begin{subfigure}[c]{0.23\linewidth}
            \centering
            \includegraphics[width=1.0\linewidth]{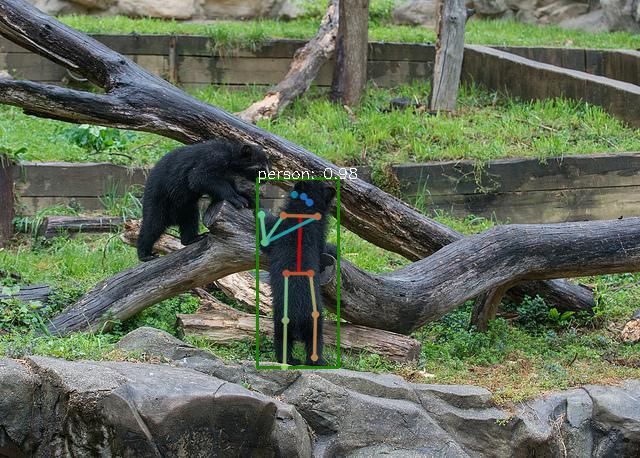}
            \vspace{-1.5em}
            \caption{Our Keypoint R-CNN distilled by HND~\cite{Matsubara19}: 1 person \\~ \label{fig:kp_hnd1}}
        \end{subfigure}
        \begin{subfigure}[c]{0.24\linewidth}
            \centering
            \includegraphics[width=1.0\linewidth]{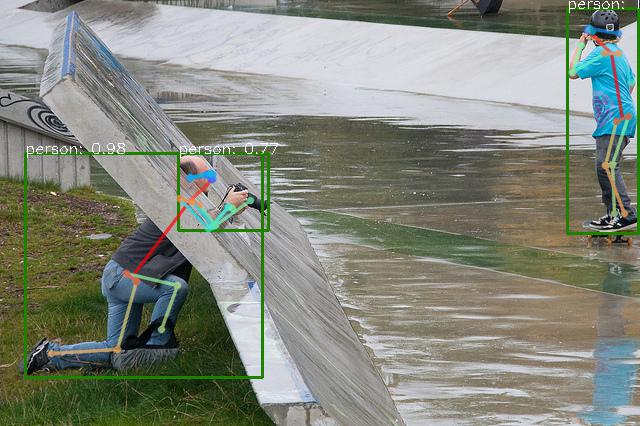}
            \vspace{-1.5em}
            \caption{Our Keypoint R-CNN distilled by HND~\cite{Matsubara19}: 3 persons \\~ \label{fig:kp_hnd2}}
        \end{subfigure}
        
        \begin{subfigure}[c]{0.24\linewidth}
            \centering
            \includegraphics[width=1.0\linewidth]{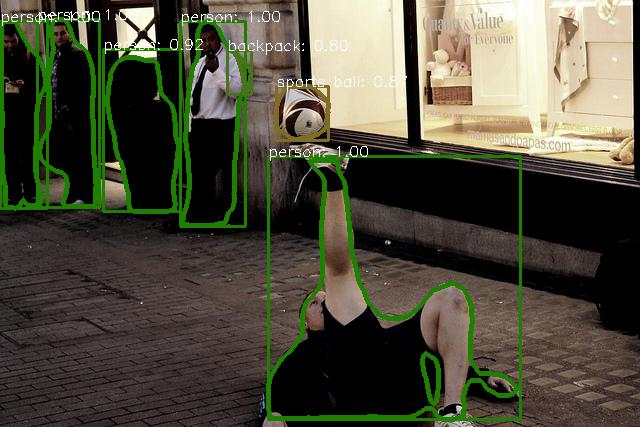}
            \vspace{-1.5em}
            \caption{{\bf Our Mask R-CNN in this work}: 5 persons, 1 sports ball and 1 backpack \label{fig:mask_ghnd1}}
        \end{subfigure}
        \begin{subfigure}[c]{0.24\linewidth}
            \centering
            \includegraphics[width=1.0\linewidth]{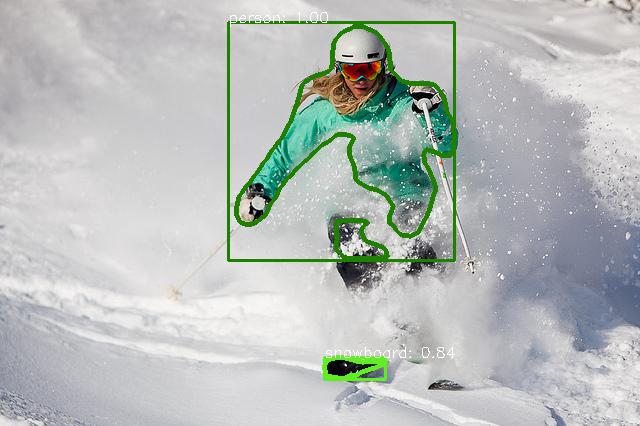}
            \vspace{-1.5em}
            \caption{{\bf Our Mask R-CNN in this work}: 1 person and 1 snowboard \\~ \label{fig:mask_ghnd2}}
        \end{subfigure}
        \begin{subfigure}[c]{0.23\linewidth}
            \centering
            \includegraphics[width=1.0\linewidth]{fig/pic/keypoint_rcnn/ghnd_000000079144.jpg}
            \vspace{-1.5em}
            \caption{{\bf Our Keypoint R-CNN in this work}: No object of interests \\~ \label{fig:kp_ghnd1}}
        \end{subfigure}
        \begin{subfigure}[c]{0.24\linewidth}
            \centering
            \includegraphics[width=1.0\linewidth]{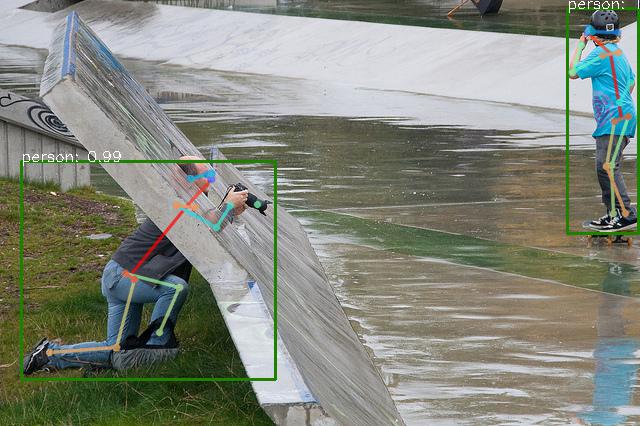}
            \vspace{-1.5em}
            \caption{{\bf Our Keypoint R-CNN in this work}: 2 persons \\~ \label{fig:kp_ghnd2}}
        \end{subfigure}
    \end{center}
    \vspace{-1em}
    \caption{Qualitative analysis. All figures are best viewed in pdf.}
    \label{fig:qualitative_analysis}
\end{figure*}

Figure~\ref{fig:qualitative_analysis} shows sampled input and output images from Mask and Keypoint R-CNNs.
Comparing to the outputs of the original models (Figs.~\ref{fig:mask_org1} - \ref{fig:kp_org2}), our Mask and Keypoint R-CNN detectors distilled by the original head network distillation (HND)~\cite{Matsubara19} suffer from false positives and negatives shown in Figs.~\ref{fig:mask_hnd1} - \ref{fig:kp_hnd1}.
As for those distilled by our generalized head network distillation, their detection performance look qualitatively comparable to the original models.
In our examples, the only significant difference between outputs of the original models and ours is that a small cell phone hold by a white-shirt man that is not detected by our Mask R-CNN shown in Fig.~\ref{fig:mask_ghnd1}.

% \begin{table}[tb]
% \caption{Hardware specifications}
% \vspace{-0.5em}
% \begin{center}
% \bgroup
% \setlength{\tabcolsep}{0.3em}
% \def\arraystretch{1.1}
% \footnotesize
% 	\begin{tabular}{llr} \toprule
% 	\multicolumn{1}{c}{Computer} & \multicolumn{1}{c}{Processor} & \multicolumn{1}{c}{RAM} \\ \midrule
% 	\shortstack[l]{Jetson TX2\vspace{0.5em}} & \shortstack[l]{ARM Cortex-A57 (4 cores)\\\hspace{0.1em} + NVIDIA Denver2 (2 cores)} & \shortstack[r]{8GB\vspace{0.5em}} \\
% 	\shortstack[l]{Desktop\vspace{0.5em}} & \shortstack[l]{Intel Core i7-6700K (4 cores)\\\hspace{0.1em} + NVIDIA GeForce RTX 2080 Ti} & \shortstack[r]{32GB\vspace{0.5em}} \\
% 	\bottomrule
% 	\end{tabular}
% \egroup
% \end{center}
% \label{table:hardware}
% \end{table}

% \begin{figure}[tb]
%     \centering
%     \includegraphics[width=1.0\linewidth]{fig/result/inference_time/faster_rcnn_inference_time.png}
%     \caption{Component-wise processing time of Faster R-CNN for different data rates.}
%     \label{fig:faster_rcnn_inference_time}
% \end{figure}

% \begin{figure}[tb]
%     \centering
%     \includegraphics[width=0.9\linewidth]{fig/result/inference_time/mask_rcnn_inference_time.png}
%     \caption{Component-wise processing time of Mask R-CNN for different data rates.}
%     \label{fig:mask_rcnn_inference_time}
% \end{figure}

\section{Related Work}
\label{appendix:related}

\vspace{1mm}
\noindent
{\bf Knowledge Distillation -} 
There are several studies discussing knowledge distillation techniques for object detectors~\cite{Li17,Chen17,Wang19,Chen19}.
However, their methods are designed to train smaller, lightweight object detectors which are not small enough to be deployed on mobile devices with limited computing capacity as shown in Tables I and II in the paper.
Thus, such approaches would not be suitable for our problem setting.

In terms of design of training loss function, FitNets~\cite{romero2014fitnets} would be the most related approach.
FitNets is a stage-wise distillation method, and has two stages for training: ``hint training'' and knowledge distillation.
Compared to FitNets, our generalized head network distillation has three advantages: (\emph{i}) it does not require a regressor which FitNets requires only in training session for adjusting their ``hint'' layer's output shape to match that of the ``guided'' layer, (\emph{ii}) our loss function is more generalized and allows us to take care of outputs from multiple layers for training, that is shown critical for detection tasks in this study, and (\emph{iii}) since the tail portion of the student model is identical to that of the pretrained teacher model, our approach is a one-stage training, thus can save training time.

\vspace{1mm}
\noindent
{\bf Edge-Assisted Video Analytics -} 
As mentioned in the introduction, the framework proposed in~\cite{Gruteser19} provides one of the most complete examples of edge-assisted vision applications.
In the paper, an augmented reality application is considered, where the mobile device performs object tracking assisted by an edge server executing object detection. In order to reduce the amount of data transported to the edge server while preserving detection accuracy, the mobile device increases the encoding quality in regions where objects were detected in previous frames. The wireless link to the edge server is assumed stable and with relatively high capacity. We observe that the data reduction approach taken in this paper may reduce the quality of detection when new objects frequently enter the vision range, as the picture quality in those regions may have been reduced.
O'Gorman and Wang~\cite{ogorman2018balancing} discuss placement decision of methods between mobile and cloud computers for balancing processing time and bandwidth for a video analytics application.

\vspace{1mm}
\noindent
{\bf Compression for Inference -}~ Xie and Kim~\cite{Xie19} present an interesting modification of JPEG to match the needs of DNNs, rather than human perception. Yet, the approach still relies on traditional wavelet compression while our approach is, in essence, forcing the compressor to extract relevant features toward the final detection objective.
Galteri \emph{et al.} \cite{galteri2018video} propose an adaptive video coding method using a learned saliency, and show the proposed method outperforms standard H.265 in terms of speed and coding efficiency through experiments on YouTube Objects dataset.

\vspace{1mm}
\noindent
{\bf DNN Splitting -} Recent literature~\cite{Kang17,Jeong18,Li18,Teerapittayanon17} proposes to split DNN models and allocate the head and tail portions to the mobile device and edge server, respectively. Thus, instead of the original sensor data, the output of the last layer of the head model is transported over the wireless channel. The problem of this approach is that many DNN models, for instance those for image processing, concentrate most of the complexity at the early layers, while not providing any significant compression until late layers. Splitting the model as is, then, allocates exceeding complexity at the weakest computing platform, while not solving the capacity problem. In fact, in~\cite{Kang17} it is shown that the optimal splitting point in most DNN models is actually pure local or edge computing depending on the channel capacity and relative devices' computing power. 

Following the work of Kang \emph{et al.}~\cite{Kang17}, some of the recent contributions propose DNN splitting methods that alter the network architectures~\cite{emmons2019cracking,eshratifar2019bottlenet,Matsubara19,emmons2019cracking,hu2020fast,shao2020bottlenet++}.
These studies, however, have the following issues: (i) lack of motivation to split the models as the size of the input data is exceedingly small, \emph{e.g.}, 32 $\times$ 32 pixels RGB images in~\cite{Teerapittayanon17,hu2020fast,shao2020bottlenet++}, (ii) consider models and network conditions specifically selected in which their proposed method is advantageous~\cite{Li18}, and/or (iii) proposed models assessed in simple classification tasks such as miniImageNet, Caltech 101, CIFAR -10, and -100 datasets~\cite{eshratifar2019bottlenet,Matsubara19,hu2020fast,shao2020bottlenet++}.

The interesting approach in~\cite{emmons2019cracking} combines DNN splitting and compression in image classification applications. The core idea is to use entropy coding applied on a transformed version of the head network model's output. The workshop paper does not discuss the resulting accuracy, and does not provide sufficient details to reproduce the results in the context considered in this paper. However, from the results it appears that this approach to compression is mostly effective when the splitting point is positioned at intermediate layers rather than early layers.
High performance object detection models forward to the final detector the output of different sections of the backbone. As a consequence, compression should be applied to all these outputs up to the splitting point, thus significantly reducing the gain. The approach proposed herein achieves compression in the early layers of the backbone network to avoid this issue.

The head network distillation technique proposed in~\cite{Matsubara19} uses distillation and bottleneck injection to boost the performance of splitting in DNNs for image classification.
As discussed throughout the paper, the different, and more complex structure, of object detection models leads to a significantly different approach to the application of distillation and bottleneck injection. Moreover, the different nature of the analysis allowed us to develop the prefiltering classifier to eliminate frames not containing objects of interest. 
\end{document}